\documentclass[conference]{IEEEtran}
\IEEEoverridecommandlockouts
\usepackage{cite}
\usepackage{amsmath,amssymb,amsfonts}
\usepackage{algorithmic}
\usepackage{graphicx}
\usepackage{textcomp}
\usepackage{xcolor}
\usepackage{algorithm}
\usepackage{algorithmic}
\usepackage{tabularx}
\usepackage{amssymb}
\usepackage{amsmath}
\usepackage{multirow}
\usepackage{booktabs} % 导入booktabs宏包，用于创建漂亮的表格线
\usepackage{newfloat}
\usepackage{listings}
\def\BibTeX{{\rm B\kern-.05em{\sc i\kern-.025em b}\kern-.08em
    T\kern-.1667em\lower.7ex\hbox{E}\kern-.125emX}}
\begin{document}

\title{PRENet: A Plane-Fit Redundancy Encoding Point Cloud Sequence Network for Real-Time 3D Action Recognition\\
% {\footnotesize \textsuperscript{*}Note: Sub-titles are not captured in Xplore and
% should not be used}
% \thanks{Identify applicable funding agency here. If none, delete this.}
}

\author{Shenglin He$^{\dag\ddag}$, Xiaoyang Qu$^{\ddag*}$, Jiguang Wan$^{\dag}$, Guokuan Li$^{\dag}$, Changsheng Xie$^{\dag}$, Jianzong Wang$^{\ddag}$  \thanks{* Corresponding author: Xiaoyang Qu (E-mail: quxiaoy@gmail.com).}\\
$^{\dag}$ Huazhong University of Science and Technology, China\\ $^{\ddag}$ Ping An Technology (Shenzhen) Co., Ltd.
}
% \IEEEauthorblockA{\textit{HuaZhong University of Science and Technology} \\
% \textit{Ping An Technology (Shenzhen) Co, Ltd.}\\
% Wuhan, China \\
% hnuhslin@163.com}
% % \and
% % \IEEEauthorblockN{2\textsuperscript{nd} Given Name Surname}
% % \IEEEauthorblockA{\textit{dept. name of organization (of Aff.)} \\
% % \textit{name of organization (of Aff.)}\\
% % City, Country \\
% % email address or ORCID}
% % \and
% % \IEEEauthorblockN{3\textsuperscript{rd} Given Name Surname}
% % \IEEEauthorblockA{\textit{dept. name of organization (of Aff.)} \\
% % \textit{name of organization (of Aff.)}\\
% % City, Country \\
% % email address or ORCID}
% % \and
% % \IEEEauthorblockN{4\textsuperscript{th} Given Name Surname}
% % \IEEEauthorblockA{\textit{dept. name of organization (of Aff.)} \\
% % \textit{name of organization (of Aff.)}\\
% % City, Country \\
% % email address or ORCID}
% % \and
% % \IEEEauthorblockN{5\textsuperscript{th} Given Name Surname}
% % \IEEEauthorblockA{\textit{dept. name of organization (of Aff.)} \\
% % \textit{name of organization (of Aff.)}\\
% % City, Country \\
% % email address or ORCID}
% % \and
% % \IEEEauthorblockN{6\textsuperscript{th} Given Name Surname}
% % \IEEEauthorblockA{\textit{dept. name of organization (of Aff.)} \\
% % \textit{name of organization (of Aff.)}\\
% % City, Country \\
% % email address or ORCID}
% }

\maketitle

\begin{abstract}
Recognizing human actions from point cloud sequence has attracted
tremendous attention from both academia and industry due to its wide applications. However, most previous studies on point cloud action recognition typically require complex networks to extract intra-frame spatial features and inter-frame temporal features, resulting in an excessive number of redundant computations. This leads to high latency, rendering them impractical for real-world applications. To address this problem, we propose a Plane-Fit Redundancy Encoding point cloud sequence network named PRENet. The primary concept of our approach involves the utilization of plane fitting to mitigate spatial redundancy within the sequence, concurrently encoding the temporal redundancy of the entire sequence to minimize redundant computations. Specifically, our network comprises two principal modules: a Plane-Fit Embedding module and a Spatio-Temporal Consistency Encoding module. The Plane-Fit Embedding module capitalizes on the observation that successive point cloud frames exhibit unique geometric features in physical space, allowing for the reuse of spatially encoded data for temporal stream encoding. The Spatio-Temporal Consistency Encoding module amalgamates the temporal structure of the temporally redundant part with its corresponding spatial arrangement, thereby enhancing recognition accuracy. We have done numerous experiments to verify the effectiveness of our network. The experimental results demonstrate that our method achieves almost identical recognition accuracy while being nearly four times faster than other state-of-the-art methods.
\end{abstract}


\section{Introduction}

Compared with RGBD videos, point cloud sequences have richer spatial and temporal contextual information for action recognition and do not involve user privacy. Therefore, point cloud sequence data has broader application value in industrial manufacturing, autonomous driving, and other fields, which is increasingly receiving attention from researchers in academia and industry. With the development of deep learning technology, point cloud 3D human action recognition has recently made breakthrough progress. However, the redundant computing introduced by previous research has resulted in high computational costs for point cloud action recognition, making it difficult to put into practical applications.
% \begin{figure}[htbp]
%     \centering
%     \includegraphics[width=0.95\linewidth]{AAAI-24.pdf}
%     \caption{Comparison of two methods for eliminating redundancy in distance error distribution on MSR-Action3D.}
%     \label{fig1}
% \end{figure}
In order to recognize human action from point cloud videos, current methods can be categorized into two main classes: 1) Raw point-based methods\cite{liu2019meteornet,DBLP:conf/iclr/FanYDYK21,Li2021RealTime3H,fan2021point}, and 2) Representation-based methods\cite{wang20203dv}. The first type of method typically uses group layers\cite{qi2017pointnet++} to model the raw point cloud to extract contextual spatio-temporal features. The second type of the method converts point cloud sequences into ordered voxels and then applies traditional grid-based convolution to these voxels. However, both of these methods have certain issues. The first approach fails to address the spatial redundancy within individual frames of point clouds, leading to a significant amount of redundant computations. The second method converts the original point cloud into an alternative form, which to some extent reduces spatial redundancy, but introduces quantization errors in the representation process, which affects the final recognition accuracy.
% Most of the current point cloud sequence-based 3D human action recognition methods can be classified into the following three categories: 1) voxel-based \cite{wang20203dv}, 2) pointnet-based\cite{qi2017pointnet, liu2019meteornet, fan2022pstnet,li2021sequentialpointnet}and 3) Transformer-based\cite{fan2021point, chen2022maple}. The first method\cite{wang20203dv} converts point cloud sequences into ordered voxels and then applies traditional grid-based convolution to these voxels. However, due to the fact that point clouds are usually sparse, directly using convolution can lead to low computational efficiency and lost too much spatial information.
% The second type of method utilizes pointnet-based\cite{qi2017pointnet,qi2017pointnet++} models to extract contextual spatio-temporal features of point cloud sequences. This method typically uses group layers\cite{qi2017pointnet++} to model the raw point cloud and then tracks the corresponding points in the sequence to preserve the temporal structure. Fan et al. proposed the third Transformer-based method. This type of method uses a 4D convolution to encode the spatio-temporal structure of the raw point cloud to avoid tracking points.  

And neither of these methods considers the time redundancy present in point cloud sequences. Many unneeded computations are repeated, resulting in the recognition task failing to achieve real-time efficiency. For example, when a person raises his right hand, his left hand and most of his body remain stationary. Since the features are extracted from each frame individually, the feature of the stationary point cloud is repeatedly extracted. It's necessary to eliminate these redundancies to achieve real-time recognition efficiency. There are still some challenges in eliminating spatial and temporal redundancy in point cloud sequences.
\begin{figure*}[t]
    \centering
    \includegraphics[width=\textwidth]{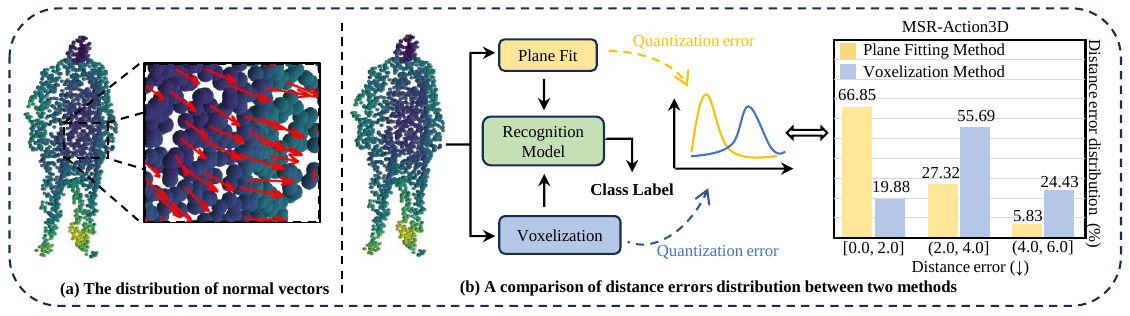}
    \caption{(a) In human action point clouds, the normal vectors of each point remain approximately parallel to those of its adjacent points. (b) We employ two distinct methods, voxelization and plane fitting, to represent the raw point cloud data. Furthermore, we utilize distance error to assess the quantization discrepancies introduced during the representation process.}
    \label{fig1}
\end{figure*}

\noindent\textbf{Redundancy Encoding.}\quad In a point cloud sequence, the redundancy part usually appears in every frame. Considering that the temporal redundancy of each frame also includes a lot of spatial redundancy, it is important to choose a suitable method to eliminate redundancy for human action point cloud sequences. We propose a method of plane fitting and propagation named the Plane-Fit Embedding module to address this challenge. We first fit several planes in a point cloud to eliminate spatial redundancy and then propagate these planes to their adjacent frames to eliminate temporal redundancy. 

\noindent \textbf{Temporal Structure Consistency Preservation.}\quad Each frame of the point cloud sequence has its temporal dimension information, also known as temporal structure. The temporal structure records where a point cloud appears in a time series. When we directly eliminate temporal redundancy in a sequence, some of its temporal structure also disappears. So the other challenge is preserving the temporal structure of redundant parts after removing the temporal redundancy. We introduce a Spatio-Temporal Consistency Encoding (STCE) module to combine the temporal feature before eliminating temporal redundancy with the spatial feature after eliminating spatial redundancy to maintain consistency between the spatial and temporal features of point clouds.

\noindent \textbf{Why plane fit?}\quad Traditional methods for eliminating point cloud spatial redundancy can be primarily categorized into two types. One type can be referred to as "Point Cloud Simplification Methods," which mainly employ downsampling or filtering techniques to reduce the number of points. The other type of method can be termed "Representation Methods," such as voxelization, which transforms points with similar spatial features into another more advanced form to eliminate spatial redundancy. The first type of method is primarily suitable for applications in single-frame static point clouds, and it is difficult to use this approach to eliminate temporal redundancy. The second type of method requires selecting an appropriate representation approach to effectively eliminate both spatial and temporal redundancies simultaneously. As shown in Figure \ref{fig1} (a), we perform normal vector estimation on each point of the human action point cloud and observe that the normal vectors of each point are approximately parallel to those of its adjacent points. And as shown in Figure \ref{fig1} (b), the distance error generated by plane fitting is significantly much smaller than that of the voxelization method. Based on this geometric property, we adopt the plane fitting method to eliminate redundancy, representing a group of adjacent points with parallel normal vectors as a plane. This approach significantly reduces the spatial redundancy of the point cloud and leads to minimal quantization errors.

We conducted extensive experiments to evaluate our PRENet's performance on three widely-used action recognition datasets: NTU RGB+D 60\cite{shahroudy2016ntu}, NTU RGB+D 120\cite{liu2019ntu}, and MSR-Action 3D\cite{li2010action}. The experimental results show that our network has achieved SOTA performance. The contribution of this paper can be summarized as follows:

\begin{itemize}
    \item We propose a redundancy encoding action recognition framework based on plane fitting and propagation to improve the speed of point cloud sequence recognition tasks significantly.
    % Plane Fit Embedding module to utilize intra-frame spatial redundancy to eliminate inter-frame temporal redundancy, which can be executed in parallel to improve recognition speed.
    \item We propose a Plane-Fit Embedding module to eliminate redundancy and a Spatio-Temporal Consistency Encoding module to combine the temporal feature before eliminating temporal redundancy with the spatial feature after eliminating spatial redundancy.
    \item We conducted extensive experiments to evaluate our PRENet's performance on three datasets. Experimental results demonstrate that our approach is nearly four times faster than other state-of-the-art (SOTA) methods.
\end{itemize}
\section{Related work}

\noindent \textbf{Deep Learning on Point Clouds Modeling.}\quad In recent years, with the release of low-cost 3D sensor devices, more and more researchers have begun to study deep learning on static point cloud modeling due to its wide range of application scenarios, such as object classiﬁcation \cite{qi2016volumetric,DBLP:journals/corr/abs-2303-10543,wang2023ssda3d,liu2023point}, part segmentation \cite{wang2019voxsegnet,qi2017pointnet++,qi2017pointnet,tang2023deep,lu2023see,xu2023approach,li20203d}, semantic scene segmentation \cite{armeni20163d,liu20173dcnn,xu2023casfusionnet,yang2023three,chen2024gaia,chen2023detecting}, and so on. Existing static point cloud modeling methods can be divided into volumetric-based, multi view-based, and point-based methods. In order to take advantage of convolutions, the volumetric-based method \cite{wang2019voxsegnet, maturana2015voxnet, wu20153d} converts the raw point cloud into regular 3D voxels, then uses 3D convolutions on them to extract contextual information. The multi view-based \cite{wei2020view,zhang2022pointclip, zhu2023pointclip,huang2023clip2point} projects 3D point clouds to different 2D views and then extracts their features through convolution operations. Both of these methods are designed to introduce convolution operations into point cloud processing. Still, quantization errors will inevitably be introduced in converting point clouds into other representations. Since the point-based method directly processes the raw point cloud, it does not generate quantization errors. PointNet \cite{qi2017pointnet} is the first point-based network that uses several sets of shared MLPs to extract local features and uses a symmetric function (max pooling) to aggregate these features to adapt to the unstructured point cloud. Furthermore, PointNet++\cite{qi2017pointnet++} proposes a set abstraction layer based on PointNet to extract point clouds' local features hierarchically, laying a good foundation for point-based research. But these methods only consider the static point cloud and do not consider the temporal dynamics of point clouds.

\noindent \textbf{Deep Learning on Point Cloud Sequence.}\quad The point cloud sequence comprises a series of continuous point cloud frames, which introduces time dimension information compared to the static point cloud. Therefore, previous static point cloud processing methods cannot be directly used to handle point cloud sequences. MeteorNet\cite{liu2019meteornet} is the first to introduce deep learning into point cloud sequence processing, which constructed the spatial-temporal neighborhoods based on PointNet++. PSTNet\cite{DBLP:conf/iclr/FanYDYK21} decomposed the temporal and spatial information of point cloud sequences and proposed a point-based convolution operation to encode point cloud sequences. These point-based methods typically use group layers\cite{qi2017pointnet++} to model the raw point cloud and then track the corresponding points in the point cloud sequence to preserve the temporal structure. \cite{fan2021point} proposed a transformer-based network, which is the first to apply the transformer to point cloud sequence processing. In \cite{wei2022spatial}, Wei et al. also proposed a transformer-based network named Point Spatial-Temporal Transformer network, which uses a self-attention mechanism to adaptively aggregate correlated inter-frame neighbor features. However, the above methods do not consider the temporal redundancy of point cloud sequences. Instead, they extract features for all frames separately and then aggregate their temporal features, leading to many redundant calculations for the network. To solve this problem, we combine the advantages of previous studies and propose PRENet to achieve real-time performance for point cloud sequence human action recognition.

\section{Methodology}
In this section, we will first introduce the overall architecture of our network. Subsequently, we will focus on the two main modules of our method: the Plane-Fit Embedding module and the Spatio-Temporal Consistency Encoding module, explaining how they address the challenges of redundancy encoding and temporal structure preservation.
\begin{figure}[t]
    \centering
    \includegraphics[width=\linewidth]{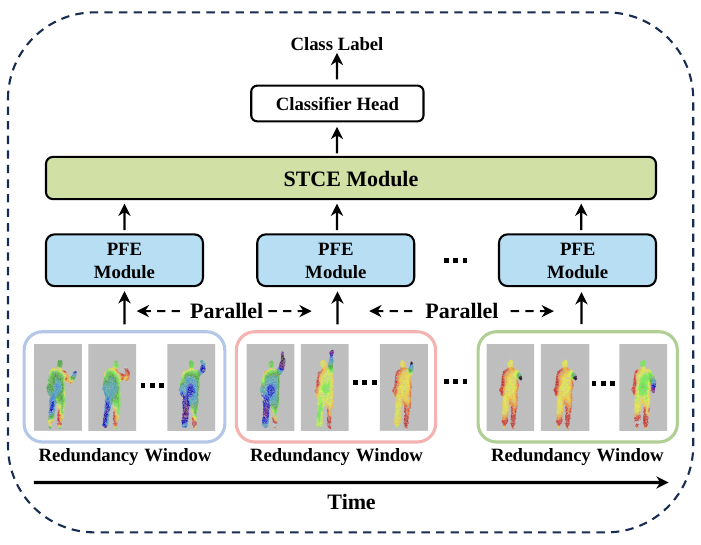}
    \caption{The overall architecture of out PRENet: PRENet consists of PFE module and STCE module. The PFE module is used to eliminate spatial and temporal redundancy in redundancy windows, while the STCE module combines the non-redundancy vectors in each window with their respective temporal structure.}
    \label{fig2}
\end{figure}
\subsection{The Overall Architecture}
The overall architecture of our PRENet is illustrated in Figure \ref{fig2}. Since only the adjacent frames in the entire point cloud sequence exhibit significant overlap in physical space, we divide the sequence into several segments, and each referred to as a redundancy window. We employ the Plane-Fit Embedding (PFE) module to eliminate spatial and temporal redundancies within each redundancy window of point cloud frames. As there is no inherent interdependency between these redundancy windows, parallel processing can be applied to enhance computational speed between windows. Subsequently, the Spatio-Temporal Consistency Encoding (STCE) module combines the non-redundant features of each redundancy window with their respective temporal structures to obtain non-redundant spatio-temporal features for the entire sequence. 

\begin{figure*}[t]
    \centering
    \includegraphics[width=1\linewidth, height=0.52\linewidth]{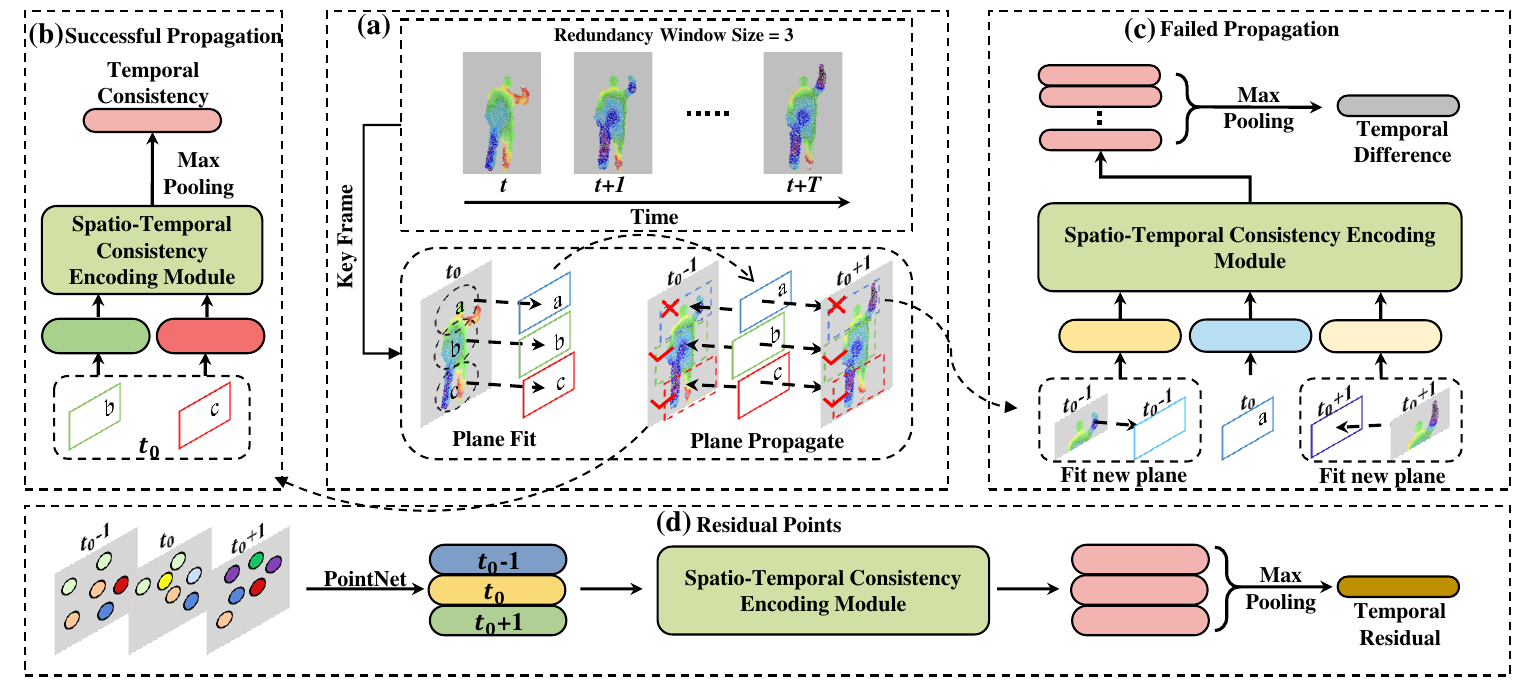}
    \caption{The architecture of the Plane-Fit Embedding module (Redundancy Window Size=3) (a) We select a key frame in each redundancy window and construct several local regions in the key frame. For each local region, we fit a plane and then propagate these planes to the other frames in the window. (b) We use a Spatio-Temporal Consistency Encoding module to combine the spatial features of these successfully propagated planes with their temporal structure and obtain their features through a max pooling operation. (c) For the failed propagated planes, we fit new planes in these local regions corresponding to other frames. (d) In each frame, some points cannot be fitted to the planes. We use the PointNet to extract features of these points and then use the STCE module to fuse their temporal information.}
    \label{fig3}
\end{figure*}

\subsection{Plane-Fit Embedding Module}
The PFE module is designed to eliminate spatial and temporal redundancies within each redundancy window to obtain non-redundant features. Its primary architecture is illustrated in Figure \ref{fig3}. Overall, the PFE module selects a frame as the key frame (typically the middle frame) in the input redundancy window, then constructs several local regions on the key frame and fits a plane for each region. And these planes are propagated to other frames in the same window through the Plane Propagate operation to eliminate temporal redundancy.

\textbf{Plane Fit.} \quad This operation is designed to eliminate spatial redundancy in each frame. Due to the unstructured and disorderly nature of point clouds, each point has many similar features and high similarity with its adjacent points. For human point clouds, this similarity is more pronounced. So every frame in the sequence has a lot of spatial redundancy that can be eliminated. The plane fitting can represent points with similar spatial features as a plane. Let $P_k^w = \left\{p_i^{(k),w}    \right\}_{i=1}^{n_k}$ denote the key frame in the $w$-th window, where $n_k$ is the number of the points in this key frame. First, we use the Farthest Point Sampling (FPS) algorithm to select $m$ points as center points and then use the KNN algorithm to select $K$ neighboring points around them to construct $m$ local regions in the key frame $P_k^w$, which the following formula can represent:
\begin{equation}\label{eq5}
    G_k^w = KNN(FPS(P_k^w))
\end{equation}
where $G_k^w \in \mathbb{R}^{m \times K \times 3}$.

In the 3D Cartesian space, a plane can be expressed as:
\begin{equation}\label{eq6}
    x + ay + bz - c = 0
\end{equation}
We can use the least squares method to fit these $m$ local regions into $m$ planes.
\begin{equation}\label{eq6}
    l_{j,k}^w = \min_{a_j,b_j,c_j}  {\textstyle \sum_{i=1}^{K}(x_{ij}+a_jy_{ij}+b_jz_{ij}-c_j)^2} , 1 \le j \le m
\end{equation}
where $l_{j,k}^w$ signifies the plane fitted through the $j$-th local region on the keyframe $k$ within the $w$-th window, and $(x_{ij},y_{ij},z_{ij})$ denotes the $i$-th point in the $j$-th local region.

A plane can be represented by at least three parameters. But in order to preserve the structural information of each local region, we also aggregate the center point of each local region into a plane. So $L_k^w = \left\{l_{1,k}^w, l_{2,k}^w,\ldots,l_{m,k}^w \right \}\in \mathbb{R}^{m \times 6} $.

\textbf{Plane Propagate.} \quad Due to the high sampling rate of modern sensors, some consecutive frames in the entire sequence may only show slight changes relative to time, resulting in very similar point clouds, which lead to temporal redundancy. It is worth mentioning that the portion with temporal redundancy within the window also contains spatial redundancy. So if we successfully propagate the plane in the key frame to other frames in the window, the spatial and temporal redundancy of the entire window will be eliminated simultaneously. After fitting $m$ planes to the key frame, we first track the center point of each plane in other frames of the current window to get $m$ redundancy neighborhoods. If these planes can represent points in redundancy neighborhoods at a certain threshold, then the propagate operation is successful, and vice versa. We will then handle the planes of successful propagation and failed propagation separately.

(1) For the successfully propagated planes, the points on those planes have almost no significant changes throughout the redundancy window. We only need one frame of the features in that part to accurately recognize human action. However, these planes have their features relative to time, so we use the STCE module to preserve the temporal structure of the plane. Then we use a max pooling layer to extract the most prominent features from these planes to enhance overall robustness. We call the final feature vector \textbf{Temporal Consistency}.

(2) For the failed propagated planes, this indicates that some points on those planes have undergone significant movement with almost no temporal redundancy. We must fit a new plane to the failed area to eliminate spatial redundancy. And since these points do not have temporal redundancy, their temporal structure throughout the entire window is crucial for correctly recognizing human action. We also use the STCE module to combine the spatial features of each frame with their temporal features. We call the final feature vector \textbf{Temporal Difference}.

(3) In redundancy windows, some points cannot be constructed as local regions or fitted by planes, and we refer to these points as residual points. For these residual points, their spatial and temporal features cannot be ignored. Therefore, as shown in Fig \ref{fig3} (d), we use PointNet\cite{qi2017pointnet} to extract the features of residual points in each frame and combine them with their temporal structure through the STCE module to obtain the final residual features, which we call \textbf{Temporal Residual}.

With the Temporal Consistency vector $TC$, Temporal Difference vector $TD$, and Temporal Residual vector $TR$, we can aggregate these three different features through a concatenation operation to obtain a non-redundancy feature for the entire redundancy window.
\begin{equation}\label{eq8}
    F_w = TC_w \oplus TD_w \oplus TR_w
\end{equation}
\begin{figure}[t]
    \centering
    \includegraphics[width=\linewidth]{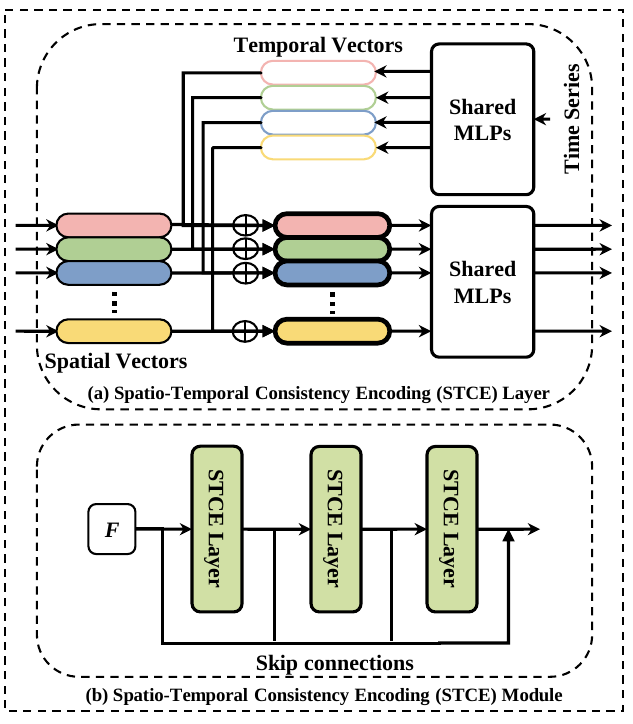}
    \caption{Spatio-Temporal Consistency Encoding module (b) mainly comprises Spatio-Temporal Consistency Encoding Layer (a). The STCE Layer is primarily composed of two shared MLPs. The STCE module stacks three STCE Layers together and incorporates skip connections to extract features of different scales.}
    \label{fig4}
\end{figure}
\subsection{Spatio-Temporal Consistency Encoding Module}
The overall architecture of the proposed Spatio-Temporal Consistency Encoding module is shown in Fig \ref{fig4}. The temporal structure of the point cloud sequence includes the order and temporal relationship of point cloud data on the timeline. For point cloud sequence, it is not enough to only consider the spatial structure of each frame. So it is imperative to maintain consistency between the spatial structure and temporal structure of each frame to accurately recognize human action. Our STCE module consists of one main component, Spatio-Temporal Consistency Encoding Layer. This section will start with STCE Layer and then transition to the STCE module.

\textbf{Spatio-Temporal Consistency Encoding Layer.} 
Our STCE Layer is primarily composed of two shared MLPs. The time sequences corresponding to each point cloud frame or redundancy window (In the PFE module, the time corresponding to each plane is used; Between redundancy windows, the time relative to each window's key frame is utilized) are encoded through a shared MLP and then added to their respective spatial vectors to align and maintain the spatio-temporal consistency. Subsequently, another set of shared MLPs is applied for further feature vector encoding.
% Our STCE Layer mainly consists of two shared MLPs, one for processing temporal features and the other for combining spatial and temporal features. For the former, we use the time series corresponding to each spatial vector as input , and the shared MLPs extract temporal features for these time series. Each spatial vector will be added to its corresponding temporal features and then fed into the second shared MLPs to obtain the spatio-temporal feature vector.

% Please add the following required packages to your document preamble:
% \usepackage{multirow}
%--------------------------------------------------------------------------
%--------------------------------------------------------------------------
%--------------------------------------------------------------------------
\begin{table*}[t]
\centering
\caption{Action recognition results on NTU RGB+D 60 and NTU RGB+D 120 datasets.}
\begin{tabularx}{\linewidth}{l *{8}{c}}
\toprule
\multirow{3}{*}{\textbf{Method}} & \multicolumn{4}{c}{\textbf{NTU RGB+D 60}} & \multicolumn{4}{c}{\textbf{NTU RGB+D 120}} \\
\cmidrule(lr){2-5} \cmidrule(lr){6-9}
& \multicolumn{2}{c}{\textbf{Accuracy (\%)}} & \multicolumn{2}{c}{\textbf{Time (ms)}} & \multicolumn{2}{c}{\textbf{Accuracy (\%)}} & \multicolumn{2}{c}{\textbf{Time (ms)}} \\
\cmidrule(lr){2-3} \cmidrule(lr){4-5} \cmidrule(lr){6-7} \cmidrule(lr){8-9}
& Cross-Subject & Cross-View & Cross-Subject & Cross-View & Cross-Subject & Cross-View & Cross-Subject & Cross-View \\
\midrule
3DV-PointNet++\cite{wang20203dv} & 88.8 & 96.3 & 308.7 & 330.5 & 82.4 & 93.5 & - & - \\
P4Transformer\cite{fan2021point} & 90.2 & 96.4 & 96.5 & 103.3 & 86.4 & 93.5 & 103.0 & 104.4 \\
PSTNet\cite{DBLP:conf/iclr/FanYDYK21} & 90.5 & 96.5 & 106.4 & 108.9& 87.0 & 93.5 & 109.3 & 112.6 \\
PSTNet++\cite{fan2021deep} & 91.4 & 96.7 & 118.2 & 120.9& 88.6 & 93.8 & 126.2 & 135.1 \\
PST-Transformer\cite{fan2022point} & 91.0 & 96.4 & 146.1 & 149.5& 87.5 & 94.0 & 155.9 & 166.8 \\
SequentialPointNet\cite{Li2021RealTime3H} & 90.3 & 97.6 & 99.8 & 102.1& 83.5 & 95.4 & 106.5 & 114.0 \\
Kinet\cite{zhong2022no} & 92.3 & 96.4 & 187.7 & 192.1& - & - & - & - \\
\midrule
PRENet(ours) & \textbf{93.2}&\textbf{97.6}& \textbf{23.2} & \textbf{25.7} & \textbf{88.9} & \textbf{96.0} & \textbf{25.4} & \textbf{27.7} \\

\bottomrule
\end{tabularx}
\label{tab1}
\end{table*}
%--------------------------------------------------------------------------
%--------------------------------------------------------------------------
%--------------------------------------------------------------------------
       
\begin{table}[t]

\centering
\caption{Action recognition results on MSR-Action 3D dataset.}
\begin{tabular}{c c c c}
\toprule
\textbf{Method} & \textbf{Frames} & \textbf{Acc (\%)} & \textbf{Time (ms)} \\
\midrule
\multirow{5}{*}{\parbox{3cm}{\centering MeteorNet \\ \cite{liu2019meteornet}}} & 4 & 78.1 & - \\
& 8 & 81.1 & - \\
& 12 & 86.5 & - \\
& 16 & 88.2 & - \\
& 24 & 88.5 & - \\
\midrule
\multirow{6}{*}{\parbox{3cm}{\centering P4Transformer \\ \cite{fan2021point}}} & 4 & 80.1 & 74.6 \\
& 8 & 83.2 & 90.1 \\
& 12 & 87.5 & 103.4 \\
& 16 & 89.6 & 128.2 \\
& 24 & 90.9 & 182.5 \\
\midrule
\multirow{6}{*}{\parbox{3cm}{\centering PST-Transformer \\ \cite{fan2021point}}} & 4 & 81.1 & 112.9 \\
& 8 & 83.9 & 136.4 \\
& 12 & 88.1 & 156.5 \\
& 16 & 91.9 & 194.1 \\
& 24 & 93.7 & 246.3 \\
\midrule
\multirow{5}{*}{\parbox{3cm}{\centering PSTNet \\ \cite{DBLP:conf/iclr/FanYDYK21}}} & 4 & 81.1 & 34.6 \\
& 8 & 83.5 & 66.8 \\
& 12 & 87.9 & 81.9 \\
& 16 & 89.9 & 167.1 \\
& 24 & 91.2 & 308.8 \\
\midrule
\multirow{5}{*}{\parbox{3cm}{\centering PSTNet++ \\ \cite{fan2021deep}}} & 4 & 81.5 & 38.4 \\
& 8 & 83.5 & 74.2 \\
& 12 & 88.1 & 90.9 \\
& 16 & 90.2 & 185.6 \\
& 24 & 92.7 & 343.0 \\
\midrule
\multirow{5}{*}{\parbox{3cm}{\centering ASTACNN \\ \cite{wang2021anchor}}} & 4 & 80.1 & - \\
& 8 & \textbf{87.5} & - \\
& 12 & \textbf{89.9} & - \\
& 16 & \textbf{91.2} & - \\
& 24 & 93.0 & - \\
\midrule
\multirow{6}{*}{\parbox{3cm}{\centering PRENet (ours)}} & 4 & \textbf{82.2} & \textbf{13.2} \\
& 8 & 85.2 & \textbf{16.5} \\
& 12 & 88.0 & \textbf{23.4} \\
& 16 & \textbf{91.2} & \textbf{28.8} \\
& 24 & \textbf{93.6} & \textbf{47.3} \\
\bottomrule
\end{tabular}
\label{tab2}
\end{table}

\textbf{Spatio-Temporal Consistency Encoding Module.} The STCE module, as shown in Fig \ref  {fig4} (b), mainly consists of three STCE layers, and skip connections are added between them to extract spatio-temporal features of different scales to improve recognition accuracy.

\section{Experiments}
In this section, we will first provide a detailed introduction to the datasets used in our experiment and various details of the experimental implementation. Secondly, we will compare our PRENet with the current SOTA methods regarding recognition accuracy and speed. Finally, at last, we conducted many ablation experiments to verify each module's effectiveness in our network.

\subsection{Dataset}
% Our experiment used three different human recognition datasets: NTU RGB+D 60\cite{shahroudy2016ntu}, NTU RGB+D 120\cite{liu2019ntu}, MSR-Action 3D\cite{li2010action}. We will introduce each of them separately.
\noindent\textbf{MSR-Action 3D:} The MSR-Action 3D dataset\cite{li2010action}, released by Microsoft Research Institute, includes 20 different action categories and ten different actors, each performing these actions in two different environments. The dataset contains 567 videos and 23k frames (270 videos for training and 297 videos for testing). The action categories included in the dataset cover common actions in daily life, such as shaking hands, waving, running, jumping, etc.

\noindent \textbf{NTU RGB+D:} NTU RGB+D 60\cite{shahroudy2016ntu} is a commonly used human action recognition dataset, including 56,880 videos composed of 60 action classes. NTU RGB+D 120\cite{liu2019ntu} is an extended version of NTU RGB+D 60 with an additional 60 extra action classes and contains 114,480 videos.

\subsection{Implementation Detail}
In this subsection, we will introduce the parameters of our network training and various other implementation details.

\noindent \textbf{Dataset preprocess:} Because all three human action recognition datasets are depth images collected by depth cameras, we must convert them into point clouds. Each depth frame is converted to a point cloud set using the public code provided by 3DV-PointNet++\cite{wang20203dv}.

\noindent \textbf{Network Details:} In the PFE module, we used three STCE modules to preserve the temporal structure of each frame. Between each parallel PFE module, one STCE module is used to aggregate the temporal structure of the entire sequence. Two sets of shared MLPs are used in each STCE module, and we will introduce the structure of each set of MLPs one by one.

\begin{itemize}
    \item In the Plane-Fit Embedding module's successful propagation part, the dimensions of these two sets of MLPs are (1, 32, 128, 6) and (6, 64, 256, 512) with $m_s$ channels, respectively, where $m_s$ is the number of planes successfully propagated.
    \item In the Plane-Fit Embedding module's failed propagation part, the dimensions of these two sets of MLPs are (1, 32, 128, 6) and (6, 64, 256, 512) with $W$ channels, respectively, where $W$ is the redundancy window size. 
    \item In the Plane-Fit Embedding module's residual points part, the dimensions of these two sets of MLPs are (1, 32, 256, 1024) and (1024, 512, 256) with $W$ channels, respectively, where $W$ is the redundancy window size. 
\end{itemize}
For other parameters in the network, we will conduct a more detailed examination in the ablation study section.

\subsection{Performance Comparison}
In this subsection, to verify the effectiveness of our proposed PRENet, we conducted performance comparison experiments on three datasets: NTU RGB+D 60\cite{shahroudy2016ntu}, NTU RGB+D 120\cite{liu2019ntu}, MSR-Action3D\cite{li2010action}. Due to the significant relationship between the overall recognition speed and the input data structure, we only compare the results with methods of the same category. Next, we will provide a detailed introduction to our experimental results.

\noindent\textbf{Comparison Results: }As shown in Table \ref{tab1}, our network has achieved the best overall recognition speed and is much faster than other SOTA methods on the large-scale dataset NTU RGB+D\cite{liu2019ntu,shahroudy2016ntu}. Our method is nearly four times faster than other state-of-the-art methods, and the recognition accuracy has not decreased significantly. We also conducted experiments on the small-scale dataset MSR-Action3D\cite{li2010action}. As shown in Table \ref{tab2}, our network demonstrated significantly faster speeds in all scenarios compared to other methods. Among them, the recognition accuracy achieved the highest results when the number of frames was 4, 16, and 24. The key to the success of our network in recognition speed is the use of plane fitting operation to simultaneously eliminate spatial and temporal redundancies in the point cloud sequence, effectively reducing a significant amount of redundant calculations at a minimal cost.

\subsection{Ablation study}
The PFE module and the STCE module are the two main components of our network. In this section, we conducted extensive ablation studies on the MSR-Action3D dataset to validate the effectiveness of these two modules.

\begin{table}[t]
\centering
\caption{The evaluation of the effectiveness of the PFE module on MSR-Action 3D dataset.}

\resizebox{\linewidth}{!}
{
\begin{tabular}{ccc}
\toprule
\textbf{Method}                   & \textbf{FLOPs (G)}                & \textbf{Inference Time (ms)} \\  \midrule
FPRENet(PointNet)        & 3.24                   & 93.6           \\
FPRENet(Transformer)     & 4.20                   & 107.5         \\
FPRENet(PSTNet)          & 3.89                   & 81.9           \\
FPRENet(MeteorNet)          & 4.11                   & 95.9           \\
\midrule
FPRENet(PFE module) & \textbf{1.84}                    & \textbf{23.4}           \\ \bottomrule 
\multicolumn{1}{l}{}     & \multicolumn{1}{l}{} &               
\end{tabular}
}

\label{tab3}
\end{table}

\subsubsection{Plane-Fit Embedding Module}
The PFE module is proposed to reduce the temporal redundancy between frames and spatial redundancy within frames in a point cloud sequence. To validate the effectiveness of this module, recognition speed and FLOPs are used as evaluation metrics. We first remove the PFE module from our PRENet network and replace it with conventional feature extraction networks such as PointNet\cite{qi2017pointnet}, Transformer\cite{vaswani2017attention}, and PSTNet\cite{DBLP:conf/iclr/FanYDYK21}. The experimental results are shown in Table \ref{tab3}. The results showed that the inference time of the network was increased by approximately 400\% after we removed the PFE module. This indicates that our PFE module has achieved good results in removing redundant calculations.

\begin{figure}[t]
    \centering
    \includegraphics[]{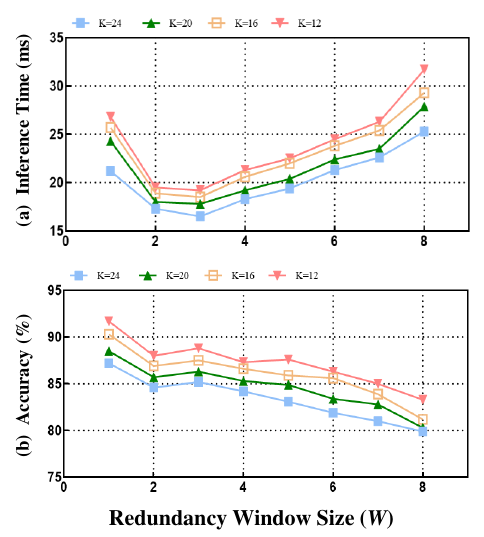}
    \caption{(a) The impact of different parameter settings on the recognition speed. (b) The impact of different parameter settings on the recognition accuracy.}
    \label{fig5}
\end{figure}

We also conducted experiments on different parameter designs in the PFE module to explore their impact on recognition accuracy and inference speed. We examined the $K$ values in the KNN algorithm and different redundancy window sizes ($W$). The experimental results are shown in Fig \ref{fig5} (a) (b). As the value of $K$ increases, the number of planes fitted gradually decreases, resulting in a faster recognition speed. However, this also leads to an increase in the quantization error, which in turn causes a reduction in accuracy. When the size of the redundancy window $W$ is larger, there will be more overall changes in the redundancy window. Directly propagating the planes from the key frame will result in a reduced recognition rate due to insufficient contextual information. It is worth noting that when the value of $W$ is set to 1, the PFE module does not perform a propagation operation. Instead, it only conducts plane fitting on the points in each frame, resulting in higher latency.

\subsubsection{Spatio-Temporal Consistency Encoding Module}
The STCE module is proposed to preserve the temporal features of each redundancy window and combine them with their spatial features to improve recognition accuracy. To verify the effectiveness of this module, we remove and replace the STCE module from our network with other networks with time modeling capabilities, such as LSTM\cite{hochreiter1997long}, RNN, GRU\cite{DBLP:conf/emnlp/ChoMGBBSB14} and Transformer\cite{vaswani2017attention}. As shown in Table \ref{tab4}, our STCE module achieved the best results in four different structures of PRENet. 

\begin{table}[t]
\centering
\caption{The evaluation of the effectiveness of the STCE module on MSR-Action 3D dataset.}
\resizebox{\linewidth}{!}
{
\begin{tabular}{ccc}
\toprule 
\textbf{Method}                    & \textbf{Accuracy (\%)} & \textbf{Inference Time (ms)} \\  \midrule
PRENet(LSTM)             & 86.39   & 39.7     \\
PRENet(RNN)              & 85.11   & 57.8     \\
PRENet(GRU)              & 83.28   & 40.4     \\
PRENet(Transformer)              & 89.68   & 78.4     \\\midrule
PRENet(STCE module) & \textbf{90.14}  & \textbf{23.4}    \\  \bottomrule
\end{tabular}

}

\label{tab4}
\end{table}
\section{Conclusion}
In this paper, we have proposed a plane-fit redundancy encoding point cloud sequence network named PRFNet for real-time 3D human action recognition. This network significantly improves recognition speed by reducing spatial and temporal redundancy in the point cloud sequence. PRENet consists of two main modules: the Plane-Fit Embedding (PFE) module and the Spatio-Temporal Consistency Encoding (STCE) module. The PFE module is responsible for eliminating spatial and temporal redundancies within the point cloud sequence. The STCE module aligns spatial and temporal features, enhancing recognition accuracy by maintaining consistency achieved through redundancy removal in each point cloud frame. The experimental results show that our network has achieved state-of-the-art performance.
\section{ACKNOWLEDGMENT}

This work is supported by the Key Research and Development Program of Guangdong Province (grant No. 2021B0101400003) and Corresponding author is Xiaoyang Qu (quxiaoy@gmail.com).

% \begin{thebibliography}{00}
% \bibitem{b1} G. Eason, B. Noble, and I. N. Sneddon, ``On certain integrals of Lipschitz-Hankel type involving products of Bessel functions,'' Phil. Trans. Roy. Soc. London, vol. A247, pp. 529--551, April 1955.
% \bibitem{b2} J. Clerk Maxwell, A Treatise on Electricity and Magnetism, 3rd ed., vol. 2. Oxford: Clarendon, 1892, pp.68--73.
% \bibitem{b3} I. S. Jacobs and C. P. Bean, ``Fine particles, thin films and exchange anisotropy,'' in Magnetism, vol. III, G. T. Rado and H. Suhl, Eds. New York: Academic, 1963, pp. 271--350.
% \bibitem{b4} K. Elissa, ``Title of paper if known,'' unpublished.
% \bibitem{b5} R. Nicole, ``Title of paper with only first word capitalized,'' J. Name Stand. Abbrev., in press.
% \bibitem{b6} Y. Yorozu, M. Hirano, K. Oka, and Y. Tagawa, ``Electron spectroscopy studies on magneto-optical media and plastic substrate interface,'' IEEE Transl. J. Magn. Japan, vol. 2, pp. 740--741, August 1987 [Digests 9th Annual Conf. Magnetics Japan, p. 301, 1982].
% \bibitem{b7} M. Young, The Technical Writer's Handbook. Mill Valley, CA: University Science, 1989.
% \end{thebibliography}
% \vspace{12pt}
% \color{red}
% IEEE conference templates contain guidance text for composing and formatting conference papers. Please ensure that all template text is removed from your conference paper prior to submission to the conference. Failure to remove the template text from your paper may result in your paper not being published.
\bibliographystyle{ieeetr}
\bibliography{reference}
\end{document}